\documentclass[10pt,twocolumn,letterpaper]{article}

\usepackage{3dv}
\usepackage{times}
\usepackage{epsfig}
\usepackage{graphicx}
\usepackage{multirow}
\usepackage{amsmath}
\usepackage{amssymb}
\usepackage{siunitx} 
\usepackage{algpseudocode}
\usepackage{algorithm}

\usepackage{times}
\usepackage{xspace}
\usepackage{xcolor}
\usepackage{booktabs }
\usepackage{enumitem}
\usepackage{makecell}

\newcommand{\figref}[1]{Fig\onedot~\ref{#1}}
\newcommand{\equref}[1]{Eq\onedot~\eqref{#1}}

\newcommand{\tabref}[1]{Tab\onedot~\ref{#1}}

\newcommand{\thickhline}{
	\noalign {\ifnum 0=`}\fi \hrule height 1pt
	\futurelet \reserved@a \@xhline
}
\DeclareRobustCommand\onedot{\futurelet\@let@token\@onedot}
\def\onedot{\ifx\@let@token.\else.\null\fi\xspace}
\def\eg{\emph{e.g.}}

\def\etc{\emph{etc}\onedot}

\DeclareRobustCommand\onedot{\futurelet\@let@token\@onedot}
\def\onedot{\ifx\@let@token.\else.\null\fi\xspace}
\def\eg{\emph{e.g.}}

\def\etc{\emph{etc}\onedot}

\newcommand{\PreserveBackslash}[1]{\let\temp=\\#1\let\\=\temp}
\newcolumntype{C}[1]{>{\PreserveBackslash\centering}p{#1}}
\newcolumntype{R}[1]{>{\PreserveBackslash\raggedleft}p{#1}}
\newcolumntype{L}[1]{>{\PreserveBackslash\raggedright}p{#1}}

\usepackage[pagebackref=true,breaklinks=true,letterpaper=true,colorlinks,bookmarks=false]{hyperref}
\usepackage{authblk}

\threedvfinalcopy 

\def\threedvPaperID{****} 

\ifthreedvfinal\fi
\begin{document}

\title{IoU Loss for 2D/3D Object Detection}
\author{Dingfu Zhou$^{1,2}$, Jin Fang$^{1,2}$,  Xibin Song$^{1,2}$, Chenye Guan$^{1,2}$, \\ Junbo Yin$^{3}$, Yuchao Dai$^{4}$ and Ruigang Yang$^{1,2}$}
\affil{$^1$Baidu Research\quad$^2$National Engineering Laboratory of Deep Learning Technology and Application, China \quad $^3$Beijing Lab of Intelligent Information Technology, School of Computer Science, Beijing Institute of Technology, China \quad $^4$ Northwestern Polytechnical University, Xi'an, China\quad \\
}

\maketitle
\begin{abstract}
In 2D/3D object detection task, Intersection-over-Union (IoU) has been widely employed as an evaluation metric to evaluate the performance of different detectors in the testing stage. However, during the training stage, the common distance loss (\eg, $L_1$ or $L_2$) is often adopted as the loss function to minimize the discrepency between the predicted and ground truth Bounding Box (Bbox). To eliminate the performance gap between training and testing, the IoU loss has been introduced for 2D object detection in \cite{yu2016unitbox} and \cite{rezatofighi2019generalized}. Unfortunately, all these approaches only work for axis-aligned 2D Bboxes, which cannot be applied for more general object detection task with rotated Bboxes. To resolve this issue, we investigate the IoU computation for two rotated Bboxes first and then implement a unified framework, IoU loss layer for both 2D and 3D object detection tasks. By integrating the implemented IoU loss into several state-of-the-art 3D object detectors, consistent improvements have been achieved for both  bird-eye-view 2D detection and point cloud 3D detection on the public KITTI \cite{geiger2012we} benchmark.   
\end{abstract}


\section{Introduction}
Object detection, as a fundamental task in computer vision and robotics, has been well studied recently. For 2D object detection, many classical frameworks have been developed, including both two-stage methods (\eg, fast R-CNN \cite{girshick2015fast}, faster R-CNN \cite{ren2015faster}) and one-stage methods (\eg, SSD \cite{liu2016ssd} and YOLO \cite{redmon2016you}). Recently, with the rapid development of the range sensors, such as the LiDAR and RGB-D cameras, 3D object detection has been attracting more and more researchers' attention. Similar with the 2D detection, some one- or two-stage based 3D object detection frameworks have been developed, such as Frustum-Pointnet \cite{qi2018frustum}, Voxel-net \cite{zhou2018voxelnet}, SECOND \cite{yan2018second}, PointPillars \cite{lang2018pointpillars} and Point R-CNN \cite{shi2019pointrcnn}. 

\begin{figure}[ht!]
	\centering
	\includegraphics[width=0.475\textwidth]{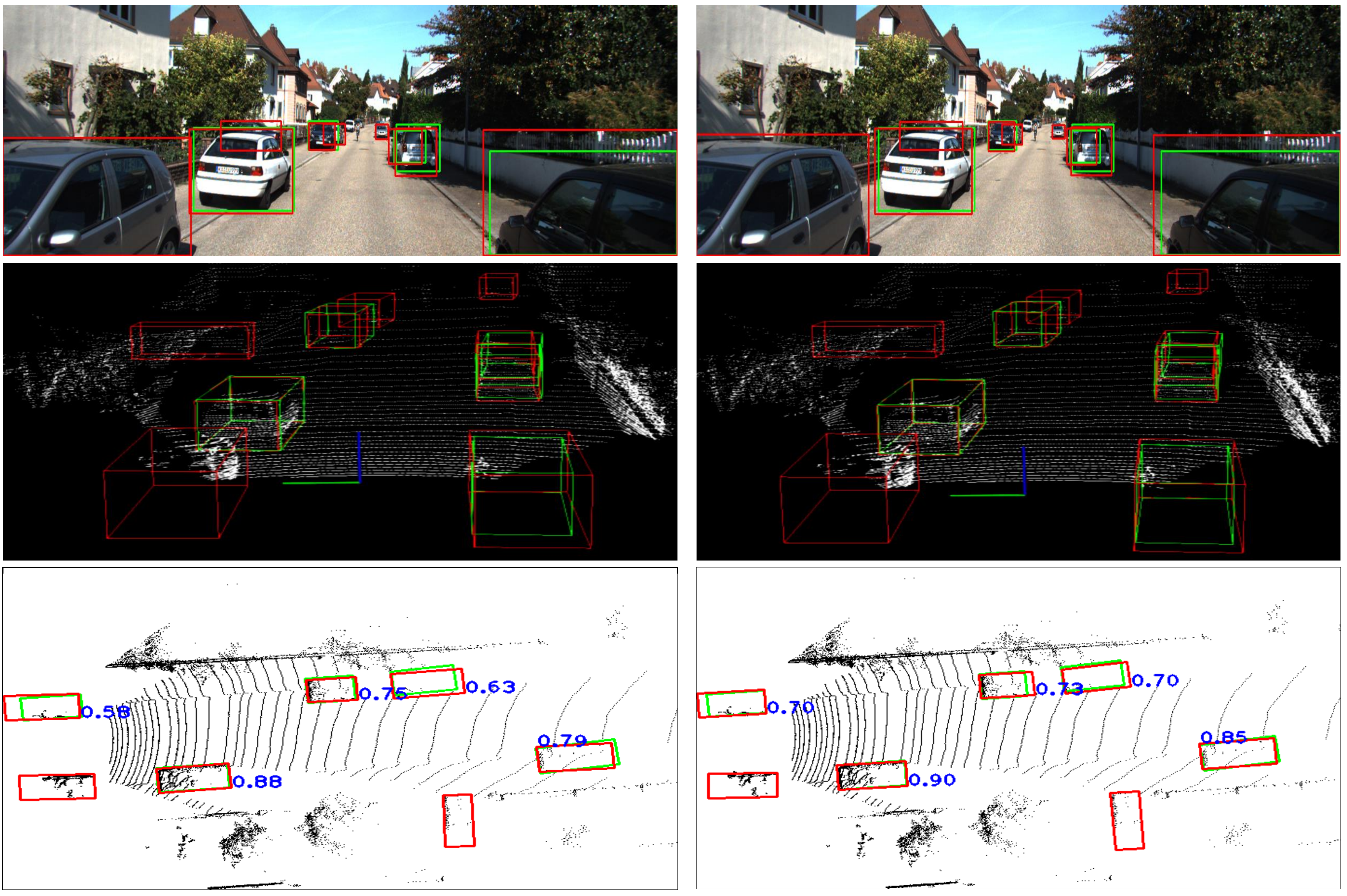}
	\caption{An example of 3D car detection results from different models trained with SECOND \cite{yan2018second} and SECOND + proposed $\mathbf{L}_{IoU}$ loss are shown in the left and right columns. The IoU value for each Bbox has been provided in the Bird-eye-view image at the bottom this figure. From this figure, we can find that the accuracy of the bounding boxes has been steadily improved by using the proposed IoU loss.}
	\label{fig:Compared_3D_detection_second_frontimage}
\end{figure}

For easy generalization, in the detection task objects are usually represented as 2D Bboxes or a 3D cuboids with several parameters, such as Bbox's center, dimension and orientation \etc. Therefore, object detection problem has been transformed as a regression task by minimizing the difference between ground truth Bbox and the predicted one. Currently, with superpower of the deep neural network, most of approaches focus on designing a better architecture backbone \cite{zhou2018voxelnet} or a better representation to extract the information of the foreground and background  objects. For the loss function, they employed the common used $L_1$ and $L_2$ distance to optimize the whole network. 

To compare the performance of different detectors, IoU metric is usually employed for evaluation, which is a total different metric compared with the $L_1$ and $L_2$  losses. As the name suggests, IoU (Intersection over Union) represents the area ratio of intersection to union of two shapes \eg Bboxes. Compared with the $L_1$ and $L_2$ distance, the IoU metric has several advantages. First, all shape properties of the Bbox has been considered in the IoU computation process, \eg, location, dimension and orientation \etc. Second, the area computation process has implicitly encoded the relationship between each parameter rather than considering them as independent variables in $L_1$ and $L_2$ loss. Finally, the IoU metric is scale invariant to the problem, which is suitable to solve the scale and range difference between each parameter.         
 
Through the analysis above, we can clearly find that there is an obvious mismatch between the objective for model training and the metric for evaluation. Frankly speaking, there is no strong correlation between the $L_1$ loss and the IoU metrics. Two predicted Bboxes may have the same $L_1$ loss with the ground truth Bbox, while the IoU value of these two Bboxes could be totally different. To eliminate this kind of gap, some efforts have been made in \cite{yu2016unitbox} and \cite{rezatofighi2019generalized} for 2D object detection. 

Unfortunately, both of them are only suitable for the easy case with axis-aligned Bboxes and none of them can be applied for the general cases with two rotated Bboxes or 3D object detection.  
In this paper, we explored the IoU calculation between two rotated Bboxes first and then implemented a unified IoU loss function which can be used for both axis-aligned and rotated 2D object detection. In addition, the new IoU loss can be also applied for 3D object detection which has only one freedom of degree for orientation. The main contribution of this paper can be summarized as:  

\begin{itemize}
	\item We investigated the IoU loss computation for two rotated 2D and 3D Bboxes;
	\item We provided a unified, framework independent, IoU loss layer for general 2D and 3D object detection tasks. 
	\item By integrating the IoU loss layer into several state-of-the-art 3D object detect frameworks such as SECOND, PointPillars and Point R-CNN, its superiority has been verified on the public KITTI 3D object detection benchmark.
\end{itemize}

\section{Related Works}

\subsection{2D object detection}
Generic object detection frameworks can mainly be divided into two directions: the first direction is also called two-stage based methods, which generate region proposals at first stage and then classify each proposal into different classes. The other one is one-stage based methods which consider the object detection as a regress and classification problem by adopting a unified framework to obtain location and classes information simultaneously. R-CNN \cite{girshick2014rich}, Fast R-CNN \cite{girshick2015fast}, Faster R-CNN \cite{ren2015faster} and Mask R-CNN \cite{he2017mask} are the most representative works of two-stages based methods, while MultiBox \cite{erhan2014scalable}, YOLO \cite{redmon2016you}, SSD \cite{liu2016ssd}, DSSD \cite{fu2017dssd} are the representative works for one-stage based methods. 

Although the design idea is slightly different between the one- and two-stage based framework, the Bbox parameters regression is a crucial component for both of them. For robust optimization and 
better regress results, different Bbox representation and loss functions have been designed. In YOLO \cite{redmon2016you}, the authors proposed to directly regress the Bbox parameters for object detection. To solve the scale sensitivity, they proposed to predict square root of the bounding
box size rather than itself. In R-CNN \cite{girshick2014rich}, the concept of prior Bbox which is also well known as proposals has been used. In this case, the Bbox regression can be transformed to predict the residual between the ground truth and the predicted Bboxes. Then $L_2$-norm is taken as the loss function for optimizing the framework. To against the outliers and noise, $L_1$-norm has been applied in Fast R-CNN \cite{girshick2015fast}. After that the $L_1$-norm has been taken as a standard loss in the object detection frameworks \cite{ren2015faster, he2017mask, liu2016ssd}.    

\subsection{3D object detection}
3D object detection in traffic scenario becomes more and more popular with the development of range sensor and the Autonomous Driving techniques. Inspired by 2D object detection, the point cloud is first 
projected into 2D (\eg bird-eye-view \cite{chen2016monocular} or front-view \cite{wu2018squeezeseg}) to obtained the 2D detection and then re-project the 2D Bbox into 3D to get the finally results. Another representative direction for 3D object detection is volumetric convolutional based methods due to the rapid development of the graphics processing resources. Voxel-net \cite{zhou2018voxelnet} is a pioneer work to detect the 3D objects directly with 3D convolutional by representing the LiDAR point cloud with voxels. For saving the GPU memory, the voxel resolution is relative large as $0.4 \text{m} \times 0.2 \text{m}  \times 0.2 \text{m} $. For each voxel, the PointNet \cite{qi2017pointnet} is applied to extract a 128-dimension features first. Based on the framework of Voxelnet, two variant methods, SECOND \cite{yan2018second} and PointPillars \cite{lang2018pointpillars} have been proposed. Different with the two directions mentioned above, PointNet \cite{qi2017pointnet} is another useful techniques for point cloud feature extraction. Along this direction, several state-of-the-art methods have been proposed for 3D object detection \cite{qi2018frustum,shi2019pointrcnn}. Similar to the 2D object detection framework, the common $L_1$-norm has been employed directly for 3D Bbox regression. 
 
\subsection{IoU Loss for Object Detection}
Most of the frameworks used a surrogate loss (\eg, $L_1$ or $L_2$ distance loss) of IoU for Bbox regression. The drawbacks of this kind of loss function have been found in ~\cite{yu2016unitbox, tychsen2018improving} and ~\cite{rezatofighi2019generalized}. In ~\cite{yu2016unitbox}, a novel IoU loss function for axis-aligned bounding box prediction has been introduced, which regresses the four bounds of a predicted box as a whole unit, performs accurate and efficient localization, shows robust to objects of varied shapes and scales, and converges fast. In \cite{tychsen2018improving}, bounded IoU loss has been developed, which is proved to be better matching the goal of IoU maximization while still providing good convergence properties. Furthermore, in ~\cite{rezatofighi2019generalized}, the authors discussed the weakness of IoU for the case of non-overlapping bounding boxes first and then introduced a generalized version of IoU (GIoU) as a new loss. Finally, the effectiveness of GIoU has been verified by integrating it into the state-of-the-art 2D object detection frameworks \cite{ren2015faster, redmon2016you, he2017mask}. All the works mentioned above target on the axis-aligned Bbox regression task, none of the works have proposed to apply the IoU loss for rotated Bbox or 3D object detection tasks.

\section{IoU for Object Detection}
IoU is also known as the Jaccard index (or the Jaccard similarity coefficient) which has been widely used to measure the similarity between finite sample sets. Generally, for two finite sample sets $\mathbf{A}$ and $\mathbf{B}$, their IoU is defined as the intersection ($\mathbf{A} \cap \mathbf{B}$) divided by the union ($\mathbf{A} \cup \mathbf{B}$) of $\mathbf{A}$ and $\mathbf{B}$. 
\begin{equation}
\textbf{IoU}(\mathbf{A},\mathbf{B}) = \frac{\mathbf{A} \cap \mathbf{B}}{\mathbf{A} \cup \mathbf{B}} = \frac{\mathbf{A} \cap \mathbf{B}}{|\mathbf{A}| + |\mathbf{B}| - \mathbf{A} \cap \mathbf{B}}
\label{eq:iou_definition}
\end{equation} 
As its definition in \cite{rezatofighi2019generalized}, {IoU} fulfills all properties of a metric, such as non-negativity, identity of indiscernibles, symmetry and triangle inequality. Especially, {IoU} is invariant to the scale which means that the similarity between two arbitrary shapes $\mathbf{A}$ and $\mathbf{B}$ is independent from the scale of their space. Due to these properties mentioned above, the {IoU} has been widely employed as evaluation metric for many task in computer vision, \eg, pixel- or instance-level image segmentation, 2D/3D object detection \etc. Particularly, we only focus on the task of object detection and its application for other tasks is beyond the scope of this paper. 

\subsection{IoU Definition for Object Detection}

For bounding box-level object detection, the target object is usually represented by a minimum Bbox rectangle in the 2D image. Base on this representation, the IoU computation between the ground bounding box $\mathbf{B}_{g}$ and the predicted bounding box $\mathbf{B}_{d}$ is defined as      
\begin{equation}
\mathbf{IoU}(\mathbf{B}_g,\textbf{B}_d) = \frac{\text{Aera of overlap } \mathbf{B}_g \text{ and } \mathbf{B}_d}{\text{Aera of union } \mathbf{B}_g \text{ and } \mathbf{B}_d}. 
\label{eq:iou_computation_2d_object}
\end{equation} 
For 3D object detection, the Bbox is simply replaced by a cuboid and the IoU value between two cuboids can be easily obtained by changing the area with volume in \equref{eq:iou_computation_2d_object}. For simplicity, we only take 2D case into the consideration here and its extension to 3D will be introduced in the following sections. 

\begin{figure}[!ht]
	\centering
	\includegraphics[width=0.175\textwidth]{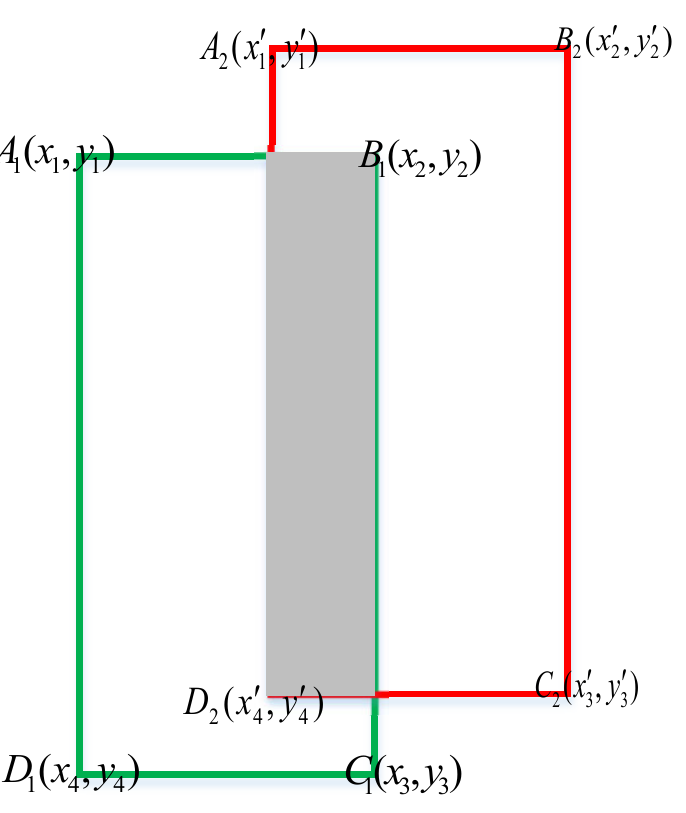}~
	\includegraphics[width=0.2\textwidth]{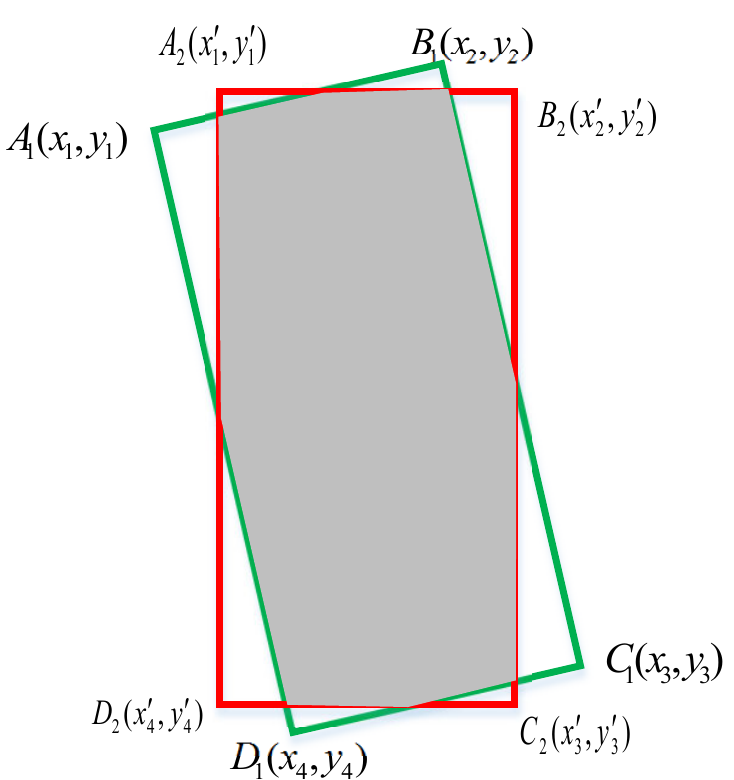}
	\caption{IoU computation for 2D: axis-aligned and rotated bounding boxes, where the green and red represent the ground truth and predicted bounding box respectively. The intersection area is highlighted in gray.}
	\label{fig:Compared_IoU_Computation}
\end{figure}

\small
\begin{algorithm}[ht!] 
	\caption{\textbf{IoU} for two axis-aligned BBoxes.}
	\begin{algorithmic}[1]
		\Require{\begin{minipage}[t]{1\textwidth}
				-Corners of the two bounding boxes:\\
				$A_1(x_1,y_1), B_1(x_2,y_1),C_1(x_2,y_2),D_1(x_1,y_2)$,\\
				$A_2(x^{'}_1,y^{'}_1), B_2(x^{'}_2,y^{'}_1),C_2(x^{'}_2,y^{'}_2),D_2(x^{'}_1,y^{'}_2)$,\\ where $x_1 \leq x_2$, $y_2 \leq y_1$ and $x^{'}_1 \leq x^{'}_2$, $y^{'}_2 \leq y^{'}_1$ 	
			\end{minipage}}
			\Ensure{
				~- $\mathbf{IoU}$ value; 
			}
			\noindent \begin{raggedright}
				\rule[0.15dd]{0.999\linewidth}{1pt}
				\par\end{raggedright}
			\State $\blacktriangleright\;$The area of $\mathbf{B}_g$: $\mathbf{Area}_{g} = (x_2 - x_1)\times (y_1 - y_2)$;
			\State $\blacktriangleright\;$The area of $\mathbf{B}_d$: $\mathbf{Area}_{d} = (x^{'}_2 - x^{'}_1)\times (y^{'}_1 - y^{'}_2)$;
			\State $\blacktriangleright\;$The area of overlap: $\mathbf{Area}_{overlap} = (\max(x_2, x^{'}_2) - \min(x_1,x^{'}_1))\times (\max(y_1,y^{'}_1) - \min(y_2, y^{'}_2))$;
			\State $\blacktriangleright\;$ $\mathbf{IoU} = \frac{\mathbf{Area}_{overlap}}{\mathbf{Area}_{g} + \mathbf{Area}_{d} - \mathbf{Area}_{overlap}}$; 
		\end{algorithmic}\label{alg:IoU_Aligned_BOXES}
	\end{algorithm}
	\normalsize
	
\subsection{Axis-aligned BBox} Usually, objects are labeled with axis-aligned BBoxes in most of the 2D object detection benchmarks, such as Pascal Visual Object Classes (VOC) Challenge \cite{everingham2010pascal}, COCO \cite{lin2014microsoft} and KITTI \cite{geiger2012we}. By taking this kind of labels as ground truth, the predicted Bboxes are also axis-aligned rectangles. For this case, the IoU computation is very easy, which can be implemented with some basic math functions, such as ``max'' and ``min'' \etc . The left of \figref{fig:Compared_IoU_Computation} illustrates an example of intersection between two axis-aligned Bboxes where the shadow area represents the intersection area. The pseudo-code of IoU computation for the axis-aligned case is given in Alg. \ref{alg:IoU_Aligned_BOXES}. 

\subsection{Rotated BBox} However, the axis-aligned box is not suitable for representing the target objects in 3D, such as the objects in the LiDAR point cloud. Usually, the 3D object is represented by a 3D cuboid. For autonomous driving scenario, the general 3D BBox with three degree-of-freedoms for rotation can be reduced to one (\eg, ``yaw'' angle) by assuming that all the objects should lay on a relative flat road ground. This kind of representation is widely used in most of the popular 3D object detection benchmarks, such as KITTI \cite{geiger2012we} and　nuScenes \cite{caesar2019nuscenes}. An example of labeled 3D object in KITTI data is given in \figref{fig:Compared_3D_detection_second_frontimage}.

For evaluation of different methods, two different strategies have been provided in KITTI: 2D Bbox overlap by projecting the 3D objects into the Bird-Eye-View (BEV) or 3D Bbox overlap directly. Here, we discussed the 2D case first and the 3D case is similar to the 2D case by adding a height dimension simply. In the BEV image, objects are represented with rotated BBoxes as described in the bottom of \figref{fig:Compared_3D_detection_second_frontimage}. The IoU computation for two rotated rectangles is more complex than axis-aligned ones because they can be intersected in many different ways. A typical example of intersection of two rotated rectangles is shown at the right of \figref{fig:Compared_IoU_Computation} and the overlap part is highlighted in blue. How to get the area of overlap part is the critical step for IoU computation. The pseudo-code for IoU computation with two rotated BBoxes is given in Alg. \ref{alg:IoU_Rotated_BOXES}.
 
\small
\begin{algorithm}[ht!] 
	\caption{\textbf{IoU} for two rotated BBoxes.}
	\begin{algorithmic}[1]
		\Require{\begin{minipage}[t]{1\textwidth}
				-Corners of the two bounding boxes:\\
			\end{minipage}}
			\Ensure{
				~- $\mathbf{IoU}$ value; 
			}
			\noindent \begin{raggedright}
				\rule[0.15dd]{0.999\linewidth}{1pt}
				\par\end{raggedright}
			\State $\blacktriangleright\;$Compute the area of $\mathbf{B}_g$: $\mathbf{Area}_{g} = \mathbf{a} \times \mathbf{b}$, where  $\mathbf{a} = \sqrt{(x_2 - x_1)^2 + (y_2 - y_1)^2}$ and $\mathbf{b} = \sqrt{(x_2 - x_3)^2 + (y_2 - y_3)^2}$;
			\State $\blacktriangleright\;$Compute the area of $\mathbf{B}_d$: $\mathbf{Area}_{d} = \mathbf{a}^{'} \times \mathbf{b}^{'}$ where $\mathbf{a}^{'} = \sqrt{(x^{'}_2 - x^{'}_1)^2 + (y^{'}_2 - y^{'}_1)^2}$ and $\mathbf{b}^{'} = \sqrt{(x^{'}_2 - x^{'}_3)^2 + (y^{'}_2 - y^{'}_3)^2}$;
			\State $\blacktriangleright\;$Determine the vertexes of overlap area if they have. 
			\State $\blacktriangleright\;$Sort these polygon vertexes in anticlockwise order; 
			\State $\blacktriangleright\;$Compute the intersection area $\mathbf{Area}_{overlap}$;
			\State $\blacktriangleright\;$ $\mathbf{IoU} = \frac{\mathbf{Area}_{overlap}}{\mathbf{Area}_{g} + \mathbf{Area}_{d} - \mathbf{Area}_{overlap}}$; 
		\end{algorithmic}\label{alg:IoU_Rotated_BOXES}
	\end{algorithm}
	\normalsize

\subsection{3D Bboxes}
As we have mentioned before, 3D object in autonomous driving is usually represented by a 3D Bbox with seven parameters, which are three for location, three for dimension and one for rotation. In this case, the IoU for two 3D Bboxes can be calculated as 
\begin{equation}
\resizebox{.9\hsize}{!}{$\mathbf{IoU}_{3D} = \frac{\mathbf{Area}_\text{overlap} \times h_\text{overlap}}{(\mathbf{Area}_{g}\times h_\text{g} + \mathbf{Area}_{d}\times h_\text{d} - \mathbf{Area}_\text{overlap}\times h_\text{overlap})}$,}
\end{equation}
where $h_\text{overlap}$ and $h_\text{union}$ represents the intersection and union in the height direction. 

\section{IoU Loss for 2D/3D BBox Regression}
So far, we have introduced IoU as a metric for two 2D and 3D BBoxes evaluation. Recently, some pioneers have succeeded in integrating the IoU loss \cite{yu2016unitbox,rezatofighi2019generalized} for BBox regression in popular 2D object detection frameworks \cite{girshick2015fast, he2017mask, redmon2016you}. Unfortunately, both of them can only handle two axis-aligned BBoxes and none of works have been proposed to deal with more general cases, such as two rotated BBoxes or 3D object detection. As we have discussed in the previous section, the computation of intersection between two rotated BBoxes is not trivial and there is not an off-the-shelf implementation in the existing deep-learning frameworks. To well rectify this situation, we first investigated the IoU loss for two rotated BBoxes and then implemented it as an unified loss layer for both 2D and 3D object detection frameworks. 

\subsection{IoU as Loss}
In \cite{yu2016unitbox} and \cite{rezatofighi2019generalized}, the effectiveness of IoU as loss function has been well proved for 2D axis aligned BBox regression task. Theoretically, it should also work well for rotated BBox because the only difference is the computation process for rotated ones is more complex than axis-aligned ones. Similar with \cite{rezatofighi2019generalized}, we defined the IoU loss as 
\begin{equation}
\mathbf{L}_{\textup{IoU}} = 1 - \mathbf{IoU}.
\label{eq:IoU_Loss}
\end{equation}
Because $\mathbf{IoU}$ satisfies $0 \leqslant \mathbf{IoU} \leqslant 1$, then the $\mathbf{L}_{\textup{IoU}}$ is also bounded between 0 and 1.   
\subsection{IoU Loss Layer}
Currently, the IoU loss for two rotated Bboxes has not implemented in any deep learning frameworks. Therefore, we implement both the forward and backward operations for this IoU loss layer.  

\subsubsection{Forward} As described in Alg. \ref{alg:IoU_Rotated_BOXES}, the forward process includes the following steps:
\vspace{-0.15cm}
\begin{enumerate}[leftmargin=0cm,itemindent=.5cm,labelwidth=\itemindent,labelsep=0cm,align=left]
	\item Compute the areas for $\mathbf{B}_d$ and $\mathbf{B}_g$, where $\mathbf{B}_d$ and $\mathbf{B}_g$ represent the predicted and ground truth BBoxes respectively; \vspace{-0.25cm}
	\item Determine the vertexes of intersection area between $\mathbf{B}_d$ and $\mathbf{B}_g$, which come from two ways: one is from the intersections of two BBoxes' edges and the other is from the BBoxes' corner who is inside the other BBox. The IoU value is zero if the vertexes don't exist. \vspace{-0.2cm}
	\item Theoretically, these vertexes form a convex hull. For computation the area of this convex hull, we need sort the vertexes in anticlockwise (or clockwise) order. First of all, the center point of these vertexes is computed.  Then, the rotation angle formed by each vertex and the center is calculated. Finally, the vertexes can be sorted by the rotation angles. \vspace{-0.2cm}
	\item Then, the intersection area is obtained by dividing it into small individual triangles. \vspace{-0.2cm}
	\item Compute the IoU value based on \equref{eq:iou_computation_2d_object} and the $\mathbf{L}_\textup{IoU}$ via \equref{eq:IoU_Loss}. 
\end{enumerate}

\subsubsection{Backward}
Currently, the derivative of common functions has been implemented in most of the public deep learning frameworks and the back-propagation process can be automatically triggered by calling these derivative computation functions. However, the analytical solution of the IoU calculation process is not easy to be provided due to the complexity of intersection between two rotated Bboxes. Especially, there exist some custom operations (intersection of two edges and sorting the vertexes \etc) whose derivative functions have not been implemented in the existing deep learning frameworks. Finally, we implement the backward operations for all these functions and we will make the source code public in the future. 

\subsubsection{Extension to GIoU Loss}
As a generalized version of IoU, GIoU has been proposed in \cite{rezatofighi2019generalized} to handle the case that two shapes don't have an intersection. In GIoU, a definition has been given to determine the distance between two non-intersected Bboxes. Generally speaking, for any two convex shapes $\mathbf{A}$, $\mathbf{B}$, a minimum area bounding shape $\mathbf{C}$ is defined as: the smallest convex shapes enclosing both $\mathbf{A}$ and $\mathbf{B}$. Usually, $\mathbf{C}$ should shares the same shape type with $\mathbf{A}$ and $\mathbf{B}$ for easy computation. Finally, the GIoU is defined as
\begin{equation}
\mathbf{GIoU} = \mathbf{IoU} - \frac{\mathbf{Area_C} - \mathbf{U}}{\mathbf{Area_C}},
\label{eq:definition_gious}
\end{equation}
where $\mathbf{U} =  \mathbf{Area_A} + \mathbf{Area_B} -\mathbf{Area}_{overlap}$. Similar as IoU loss, we also extended the GIoU loss for the case of rotated Bboxes.

\section{Experimental Results}
 \begin{table*}[ht!]
 	\centering
 	\resizebox{0.995\textwidth}{!}
 	{%
 		\begin{tabular}{lcccccccccccc}
 			\hline
 			\multicolumn{1}{l}{\multirow{2}{*}{\textbf{Loss Types}}} & \multicolumn{3}{c}{\textbf{AP70}}  & \multicolumn{3}{c}{\textbf{AP75}} 	& \multicolumn{3}{c}{\textbf{AP80}} & \multicolumn{3}{c}{\textbf{mAP}} \\
 			\multicolumn{1}{l}{} & \multicolumn{1}{l}{Easy} & \multicolumn{1}{l}{Mod} & \multicolumn{1}{l}{Hard} 
 			& \multicolumn{1}{l}{Easy} & \multicolumn{1}{l}{Mod} & \multicolumn{1}{l}{Hard} 
 			& \multicolumn{1}{l}{Easy} & \multicolumn{1}{l}{Mod} & \multicolumn{1}{l}{Hard} 
 			& \multicolumn{1}{l}{Easy} & \multicolumn{1}{l}{Mod} & \multicolumn{1}{l}{Hard} \\ \hline
 			SECOND\cite{yan2018second} + $\mathbf{L}_1$ & 88.15 & 78.33 & 77.25　& 81.37 & 66.86 & 65.48 & 59.56 & 48.90 & 44.45 & 62.41 & 57.52 & 56.23 \\ \hline
 			SECOND + $\mathbf{L}_\text{IoU}$  & \textbf{89.16} & 78.99 & 77.78 & \textbf{83.40} & \textbf{73.36} & \textbf{66.72} & \textbf{66.36} & \textbf{52.60} & \textbf{50.61} & \textbf{64.55} & \textbf{58.96} & \textbf{57.61} \\
 			Rel improvement $\Uparrow$ & 0.94\% & 0.82\% & 0.91\% & \textbf{2.49}\% & \textbf{9.72}\% & \textbf{1.89}\% & \textbf{11.42}\% & \textbf{7.57}\% & \textbf{13.86}\%& \textbf{3.43\%} & \textbf{2.50\%} & \textbf{2.45\%} \\ \hline
 			SECOND + $\mathbf{L}_\text{GIoU} $ & {89.15} & \textbf{79.14} & \textbf{78.11} & 82.56 & 72.98 & 66.34 & 64.27 & 51.67 & 50.11 & 64.38 & 58.73 & 57.20 \\
 			Rel improvement $\Uparrow$ & \textbf{1.13}\% & \textbf{1.03}\% & \textbf{1.11}\%& 1.46\% & 9.15\% & 1.31\% & 7.91\% & 5.66\% & 12.73\% & 3.16\% & 2.10\% & 1.72\% \\ \hline
 		\end{tabular}%
 	}
 	\caption{\normalfont Evaluation results by training SECOND \cite{yan2018second} with $\mathbf{L}_1$ loss and proposed losses on validation dataset of the KITTI 3D car detection benchmark. All the numbers are the higher the better. The best result of each column has been highlighted with bold font.}
 	\label{tab:Comparison_second_val_kitti}
 \end{table*}
 
  \begin{table}[ht!]
  	\centering
  	\resizebox{0.485\textwidth}{!}
  	{%
  		\begin{tabular}{lcccccccccccc}
  			\hline
  			\multicolumn{1}{l}{\multirow{2}{*}{\textbf{Loss Types}}} & \multicolumn{3}{c}{\textbf{BEV(AP70)}}  & \multicolumn{3}{c}{\textbf{mAP}} \\
  			\multicolumn{1}{l}{} & \multicolumn{1}{l}{Easy} & \multicolumn{1}{l}{Mod} & \multicolumn{1}{l}{Hard} 
  			& \multicolumn{1}{l}{Easy} & \multicolumn{1}{l}{Mod} & \multicolumn{1}{l}{Hard} \\ \hline
  			SECOND\cite{yan2018second} + $\mathbf{L}_1$ & 89.92 & 87.88 & 86.72 　& 70.63 & 67.71 & 65.82  \\ \hline
  			SECOND + $\mathbf{L}_\text{IoU}$  & 90.21 & 88.25  & 87.56 & 71.39 & 68.37 & 66.23  \\
  			Rel improvement $\Uparrow$ & 0.32\% & 0.42\% & 0.97\% & 1.07\% & 0.97\% & 0.62\%  \\ \hline
  			SECOND + $\mathbf{L}_\text{GIoU} $ & \textbf{90.25} & \textbf{88.51} & \textbf{87.65} & \textbf{73.35} & \textbf{68.48} & \textbf{66.92} \\
  			Rel improvement $\Uparrow$ & \textbf{0.37\%} & \textbf{0.72\%} & \textbf{1.07\%}& \textbf{3.85}\% & \textbf{1.14}\% & \textbf{1.67}\% \\ \hline
  		\end{tabular}%
  	}
  	\caption{\normalfont Evaluation results of SECOND \cite{yan2018second} with $\mathbf{L}_1$ and IoU losses on validation dataset of the KITTI BEV car detection benchmark. The number is the higher the better. The best result of each column is highlighted with bold font.}
  	\label{tab:Comparison_second_val_kitti_bird_eye_view}
  \end{table}

The proposed loss layer is an framework independent modular which can be integrated into any regression-based 2D or 3D object detection methods. Different with \cite{rezatofighi2019generalized}, the proposed loss layer is more general on both axis-aligned and non-axis-aligned cases, such as 2D BEV or 3D object detection. We integrate the proposed IoU/GIoU loss on different types of 3D object detection frameworks and then compare their performances on the public third-party 3D object detection benchmark.

\textbf{Baselines:} three state-of-the-art 3D object detectors have been evaluated here: SECOND\cite{yan2018second}, PointPillars \cite{lang2018pointpillars} and PointRCNN \cite{shi2019pointrcnn}. SECOND is a voxel-based one-stage object detector, which is an advance version of VoxelNet \cite{zhou2018voxelnet} by adding a sparse convolution operations implemented by themself.
PointPillars is an acceleration version of SECOND which represents the point cloud by pillars rather than voxels. First, PointNet\cite{qi2017pointnet} is employed to extract features for each ``Pillar'' and then the ``Pillar'' is taken as the minimum elements for the further convolution network. Compared with SECOND, the pillar expression is much faster than voxel representation. Different with the previous two methods, PointRCNN is a two-stage 3D object detector, which combines the point segmentation and region proposal at the first stage and the Bbox refinement is executed at the second stage of the framework.  

\textbf{Dataset:} we train all the baselines and evaluate them on KITTI \cite{geiger2012we} 3D object detection benchmark. This data has been divided into training and testing two subsets, which consists of 7481 and 7518 frames respectively. Since the ground truth for the testing set is not available, we subdivide the training data into a training and validation set as described in \cite{zhou2018voxelnet, yan2018second}. Finally, we obtained 3,712 data samples for training and 3,769 data samples for validation. On the KITTI benchmark, the objects have been categorized into ``easy'', ``moderate'' and ``hard'' based on their height in the image and occlusion ratio, etc. For each frame, both the camera image and the LiDAR point cloud has been provided, while only the point cloud has been used for our object detection here and the RGB image is only used for visualization.

\textbf{Evaluation protocol:} In this paper, we employ similar evaluation metric as KITTI \cite{geiger2012we} to report all our results. In \cite{geiger2012we}, all the objects have been divided into ``Easy'', ``Moderate'' and ``Hard'' category based on their distances and occlusion ratios. For each category, we calculate the Average precision (AP) by giving a certain IoU threshold. Different with KITTI, we set three different thresholds here. Beside this, we also give the mean Average Precision (mAP) across different value of IoU thresholds, i.e. $\textbf{IoUs} = \{0.50, 0.55,  \dots, 0.90, 0.95\}$ to evaluate the performance of detectors at different thresholds.
\begin{figure*}[!ht]
	\centering
	\includegraphics[width=0.75\textwidth]{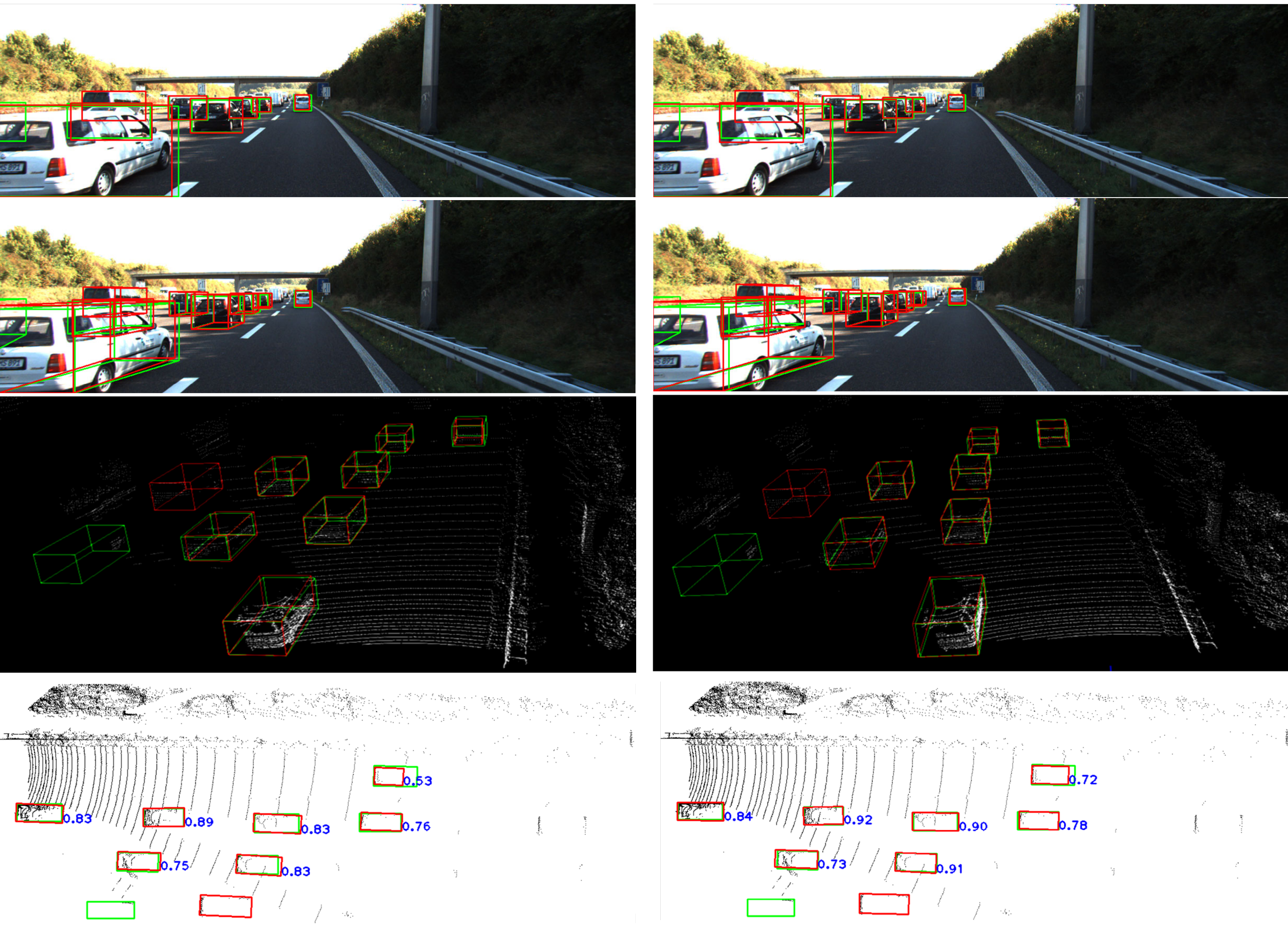}
	\caption{An example of 3D car detection results with different methods, where the left is from original SECOND method and the right is the SECOND with the proposed IoU loss.}
	\label{fig:Compared_3D_detection_second}
\end{figure*}

\begin{table*}[ht!]
	\centering
	\resizebox{0.9\textwidth}{!}
	{%
		\begin{tabular}{lcccccccccccc}
			\hline
			\multicolumn{1}{l}{\multirow{2}{*}{\textbf{Loss Types}}} & \multicolumn{3}{c}{\textbf{AP70}}  & \multicolumn{3}{c}{\textbf{AP75}} 	& \multicolumn{3}{c}{\textbf{AP80}} & \multicolumn{3}{c}{\textbf{mAP}} \\
			\multicolumn{1}{l}{} & \multicolumn{1}{l}{Easy} & \multicolumn{1}{l}{Mod} & \multicolumn{1}{l}{Hard} 
			& \multicolumn{1}{l}{Easy} & \multicolumn{1}{l}{Mod} & \multicolumn{1}{l}{Hard} 
			& \multicolumn{1}{l}{Easy} & \multicolumn{1}{l}{Mod} & \multicolumn{1}{l}{Hard} 
			& \multicolumn{1}{l}{Easy} & \multicolumn{1}{l}{Mod} & \multicolumn{1}{l}{Hard} \\ \hline
			PointPillars \cite{lang2018pointpillars} + $\mathbf{L}_1$ & 87.29 & 76.99 & 70.84 & 72.39 & 62.73 & 56.40  & 47.23 & 40.89 & 36.31 & 58.62 & 54.86 & 52.74 \\ \hline
			PointPillars　+ $\mathbf{L}_\text{IoU}$ & 87.88 & 77.92 & 75.70 &76.18 & 65.83 & 62.12  & \textbf{57.82} & \textbf{45.03} & \textbf{42.95} & \textbf{62.07} & \textbf{57.11} & \textbf{55.67} \\
			Rel improvement$\Uparrow$& 0.68\% & 1.21\% & 6.86\% & 5.24\% & 4.94\% & 10.14\% & \textbf{22.4}\% & \textbf{10.1}\% & \textbf{18.28}\% & \textbf{5.89}\% & \textbf{4.10}\% & \textbf{5.56}\% \\ \hline
			PointPillars + $\mathbf{L}_\text{GIoU}$ & \textbf{88.43} & \textbf{78.15} & \textbf{76.34}&\textbf{76.93} & \textbf{66.36} & \textbf{63.68} &  56.36 & 44.43 & 42.72& 61.94 & 56.65 & 55.13 \\
			Rel improvement$\Uparrow$& \textbf{1.34}\% & \textbf{1.47}\% & \textbf{7.62}\% & \textbf{6.27}\% & \textbf{5.78}\% & \textbf{12.9}\% & 19.3\% & 8.66\% & 17.65\% & 5.53\% & 2.44\% & 4.17\% \\ \hline
		\end{tabular}%
	}
	\caption{\normalfont Evaluation results by training PointPillar~\cite{lang2018pointpillars} with $\mathbf{L}_1$ loss and proposed losses on validation dataset of the KITTI 3D car detection benchmark. All the numbers are the higher the better. The best result of each column has been highlighted with bold font.}
	\label{tab:Comparison_pointpillars_val_kitti}
\end{table*}

\subsection{SECOND~\cite{yan2018second}}
\textbf{Training protocol:} the officially released code \footnote{\url{https://github.com/traveller59/second.pytorch}} by the authors has been used for training the baseline SECOND model. We use exactly the same config file provided by the author and follow the same training protocol to achieve the baseline results on the KITTI benchmark. Compared with the baseline network, we just simply replace the regress loss $\mathbf{L}_1$ with our self-implemented 
$\mathbf{L}_\text{IoU}$ and $\mathbf{L}_\text{GIoU}$ losses. We used nearly the same training strategy as the baseline \eg, same iteration steps and learning rate \etc. The only difference is that we decrease the threshold of an anchor be considered as a positive sample during the training from 0.6 to 0.5, which means that there are more positive anchors have been involved in the training process. We set threshold at 0.6 in baseline framework because it gives better results than 0.5.

\textbf{Results:} The comparison of the IoU losses with original SECOND method for 3D car detection on KITTI benchmark has been given in \tabref{tab:Comparison_second_val_kitti}. On this benchmark, the matching IoUs threshold ${0.7}$ is used to evaluation, however, as mentioned by \cite{tychsen2018improving}, the hyper-parameters (e.g., the matching IoUs threshold) usually have big influences on the detectors if only a certain matching IoUs threshold. Therefore, three different matching thresholds $\{ {0.70}, {0.75}, {0.80} \}$ and the mAP have been applied here for evaluation.    

From this table, we can find that the proposed $\mathbf{L}_\text{IoU}$ and $\mathbf{L}_\text{GIoU}$ gives slightly better results than baseline for all the three categories (``easy'', ``moderate'' and ``hard'') at the IoUs matching threshold $0.7$. Compared with the baseline, around 1\% relative improvement has been given by $\mathbf{L}_\text{IoU}$ and $\mathbf{L}_\text{GIoU}$ for the three categories. At this threshold, the $\mathbf{L}_\text{GIoU}$ performs slightly better than $\mathbf{L}_\text{IoU}$. 

We also find an interesting phenomenon that the $\mathbf{L}_\text{IoU}$ and $\mathbf{L}_\text{GIoU}$ losses give much more improvements than baseline when the IoUs matching threshold at a higher value. We can see clearly that the improvements at $\mathbf{AP80}$ are much greater than $\mathbf{AP70}$. At $\mathbf{AP80}$, the relative improvements for the three categories can reach \textbf{11.42}\%, \textbf{7.57}\% and \textbf{13.86}\% respectively by using the $\mathbf{L}_\text{IoU}$ loss. At this threshold, the improvements for $\mathbf{L}_\text{GIoU}$ loss can achieve to 7.91\%, 5.66\% and 12.73\% which performs slightly worse than $\mathbf{L}_\text{IoU}$.  

The mAPs for all methods have been given in the last column of this table. We can also easily find that the detection performance has been steadily improved by the $\mathbf{L}_\text{IoU}$ and $\mathbf{L}_\text{GIoU}$ losses. By using the new loss, all the detection rates have an average improvement of 2\%  and the improvement can reach 3\% for some specific category. 

 \begin{table}[ht!]
  	\centering
  	\resizebox{0.485\textwidth}{!}
  	{%
  		\begin{tabular}{lcccccccccccc}
  			\hline
  			\multicolumn{1}{l}{\multirow{2}{*}{\textbf{Loss Types}}} & \multicolumn{3}{c}{\textbf{BEV(AP70)}}  & \multicolumn{3}{c}{\textbf{mAP}} \\
  			\multicolumn{1}{l}{} & \multicolumn{1}{l}{Easy} & \multicolumn{1}{l}{Mod} & \multicolumn{1}{l}{Hard} 
  			& \multicolumn{1}{l}{Easy} & \multicolumn{1}{l}{Mod} & \multicolumn{1}{l}{Hard} \\ \hline
  			PointPillars \cite{lang2018pointpillars} + $\mathbf{L}_1$ & 90.07　&  87.06 &  83.81  & 69.11& 66.84 & 65.36  \\ \hline
  			PointPillars + $\mathbf{L}_\text{IoU}$  & 90.24& 88.02& 86.64 & 71.33& \textbf{68.11}& 66.53 \\
  			Rel improvement $\Uparrow$ & 0.19\% & 1.10\% & 3.38\% & 3.21\% & \textbf{1.90}\% & 1.79\%  \\ \hline
  			PointPillars + $\mathbf{L}_\text{GIoU} $ & \textbf{90.35}& \textbf{88.26}& \textbf{87.04} & \textbf{71.74} & 68.04 & \textbf{66.63}\\
  			Rel improvement $\Uparrow$ & \textbf{0.31\%} & \textbf{1.37\%} & \textbf{3.85\%}& \textbf{3.81}\% & 1.80\% & \textbf{1.94}\% \\ \hline
  		\end{tabular}%
  	}
  	\caption{\normalfont Evaluation results by training PointPillars \cite{lang2018pointpillars} with $\mathbf{L}_1$ loss and proposed losses on KITTI validation dataset for BEV image. The best result of each column has been highlighted with bold font.}
  	\label{tab:Comparison_pointpillars_val_kitti_bird_eye_view}
  \end{table}

The detection results of BEV image is given in \tabref{tab:Comparison_second_val_kitti_bird_eye_view}. Compared with the baseline, we can also find that the detection rate has been slightly improved with the proposed IoU loss for all the three categories. An example of detection results in BEV image and point cloud is given in Fig. \ref{fig:Compared_3D_detection_second}. The bottom of this figure gives the 2D detection in BEV image, where the number around each Bbox is the IoU value in 3D. We can found that most of the values in right is larger than left, which means that the bounding box's accuracy has been consistently improved by the proposed IoU loss.
\subsection{PointPillar~\cite{lang2018pointpillars}}

\textbf{Training protocol:} the officially released code \footnote{\url{https://github.com/nutonomy/second.pytorch}} by the authors has been used for training the PointPillars baseline model. We reproduce the baseline results on the KITTI benchmark, following the officially configure file and training protocols. Similar with SECOND method, we replaced the regression loss with our proposed $\mathbf{L}_\text{IoU}$ and $\mathbf{L}_\text{GIoU}$ losses and decrease the foreground threshold from 0.6 to 0.5. 

\textbf{Results:} The comparison of the proposed losses with original PointPillars is given in \tabref{tab:Comparison_pointpillars_val_kitti}. The similar evaluation criterion is applied here too. From \tabref{tab:Comparison_pointpillars_val_kitti}, the power of the proposed losses is demonstrated clearly. For $\mathbf{AP70}$, the detection rates have been improved around 1\% for ``easy'' and ``moderate'' categories, 6.86\% and 7.63\% for ``hard'' category with $\mathbf{L}_\text{IoU}$ and $\mathbf{L}_\text{GIoU}$ losses respectively. For $\mathbf{AP80}$, the proposed $\mathbf{L}_\text{IoU}$ loss can achieve a significant improvement by \textbf{22.4\%}, \textbf{10.1\%}, \textbf{18.28\%} on the three categories, for the proposed $\mathbf{L}_\text{GIoU}$, which performs slightly worse, but also inspiring, promoted the baseline by 19.3\%, 8.66\%, 17.65\% respectively. 

The mAP values are shown in the last column, which have been steadily improved by over 4\% roughly for both the $\mathbf{L}_\text{IoU}$ and  $\mathbf{L}_\text{GIoU}$ losses compared with the baseline. And the improvement can reach 5\% for some specific categories.The detection results on the BEV images are shown in \tabref{tab:Comparison_pointpillars_val_kitti_bird_eye_view}. From this table, we can also find steadily improvement with the proposed $\mathbf{L}_\text{IoU}$ and  $\mathbf{L}_\text{GIoU}$ losses.
\subsection{PointRCNN~\cite{shi2019pointrcnn}}
\textbf{Training protocol:} different with the previous two methods, PointRCNN is a two-stage based framework. At the first stage, RPN network is employed to generate region proposal first and the Bbox refinement is executed at the second stage to get the final results. We trained the baseline with the official released source code here \footnote{\url{https://github.com/sshaoshuai/PointRCNN}}. Currently, we kept the RPN part unchanged and integrated the proposed IoU loss only at the second stage. The loss function in the stage includes class classification loss, bin classification loss and the regression loss. Here, we keep the first two parts unchanged and replace the regression loss with the proposed $\mathbf{L}_\text{IoU}$ and $\mathbf{L}_\text{GIoU}$. To be clear, based on the officially released code, we cannot obtain the results reported in their paper. Therefore, the baseline reported here is the best model that we can achieve. Particularly, we train the baseline and the proposed losses by using the same training strategies for fair comparison.  

\begin{table*}[ht!]
	\centering
	\resizebox{0.95\textwidth}{!}
	{%
		\begin{tabular}{lcccccccccccc}
			\hline
			\multicolumn{1}{l}{\multirow{2}{*}{\textbf{Loss Types}}} & \multicolumn{3}{c}{\textbf{AP70}}  & \multicolumn{3}{c}{\textbf{AP75}} 	& \multicolumn{3}{c}{\textbf{AP80}} & \multicolumn{3}{c}{\textbf{mAP}} \\
			\multicolumn{1}{l}{} & \multicolumn{1}{l}{Easy} & \multicolumn{1}{l}{Mod} & \multicolumn{1}{l}{Hard} 
			& \multicolumn{1}{l}{Easy} & \multicolumn{1}{l}{Mod} & \multicolumn{1}{l}{Hard} 
			& \multicolumn{1}{l}{Easy} & \multicolumn{1}{l}{Mod} & \multicolumn{1}{l}{Hard} 
			& \multicolumn{1}{l}{Easy} & \multicolumn{1}{l}{Mod} & \multicolumn{1}{l}{Hard} \\ \hline
			PointRCNN \cite{shi2019pointrcnn} & 88.14　&  77.58 &  75.36  & 73.27 & 63.54 & 61.08  & 44.21 & 38.88 & 34.62 & 59.44 & 54.35 & 52.79 \\ \hline
			PointRCNN　+ $\mathbf{L}_\text{IoU}$ & 88.83 & 78.80 & \textbf{78.18} & 77.42 & 67.83 & 66.85  & {58.22} & {49.09} & {45.38} & \textbf{63.47} &57.71 & 56.67 \\
			Rel improvement$\Uparrow$& 0.78\% & 1.57\% & \textbf{3.74}\% & 5.66\% & 6.75\% & 9.44\% & {31.6}\% & {26.26}\% & 31.08\% &  \textbf{6.78}\% & 6.18\% &7.35\% \\ \hline
			PointRCNN + $\mathbf{L}_\text{GIoU}$ & \textbf{88.84}& \textbf{78.85}& 78.15&\textbf{77.47} & \textbf{67.98} & \textbf{67.18} & \textbf{59.80} & \textbf{51.25} & \textbf{46.50} & 63.12 & \textbf{57.96}& \textbf{56.92} \\
			Rel improvement$\Uparrow$& \textbf{0.79\%} & \textbf{1.64\%} & 3.70\% & \textbf{5.73}\% & \textbf{6.99}\% & \textbf{9.99}\% & \textbf{35.3\%} & \textbf{31.81\%} & \textbf{34.31\%} & 6.19\% & \textbf{6.64}\% & \textbf{7.82}\% \\ \hline
		\end{tabular}%
	}
	\caption{\normalfont Evaluation results by training PointRCNN \cite{shi2019pointrcnn} with $\mathbf{L}_1$ loss and proposed losses on KITTI validation dataset for BEV image. The best result of each column has been highlighted with bold font.}
  	\label{tab:Comparison_PointRCNN_val_kitti_3D}
\end{table*}
\textbf{Results:} the comparison of PointRCNN with different losses is given in \tabref{tab:Comparison_PointRCNN_val_kitti_3D}. Similar to the previous methods, we can easily find that both the $\mathbf{L}_\text{IoU}$ and $\mathbf{L}_\text{GIoU}$ can improve baseline's performance at different IoU threshold for all the categories. Especially, the detection rates have been improved by a big margin when we have a higher IoU threshold. Furthermore, for the mAP criterion, both the $\mathbf{L}_\text{IoU}$ and $\mathbf{L}_\text{GIoU}$ also give a big improvement compared with the original PointRCNN. Based on this experiment, we can conclude that the proposed IoU loss can also work for two-stage based method. 

\subsection{Comparison with Other Methods}
\begin{table}[]
	\resizebox{0.45\textwidth}{!}
	{%
		\begin{tabular}{r c ccc }
			\hline
			\multicolumn{1}{c }{\multirow{2}{*}{Methods}}& \multicolumn{1}{c}{\multirow{2}{*}{Modality}} & \multicolumn{3}{c}{\textbf{AP70}} \\
			\multicolumn{1}{c}{}& \multicolumn{1}{c}{} & \multicolumn{1}{l}{Easy} & \multicolumn{1}{l}{Mod} & \multicolumn{1}{l}{Hard}\\ \hline
			MV3D\cite{chen2017multi} &LiDAR+Mono& 71.29 & 62.68 & 56.56 \\
			F-PointNet\cite{qi2018frustum} &LiDAR+Mono& 83.76 & 70.92 & 63.65 \\
			AVOD-FPN\cite{ku2018joint} &LiDAR+Mono& 84.41 & 74.44 & 68.65 \\
			ContFusion\cite{liang2018deep} &LiDAR+Mono& 86.33 & 73.25 & 67.81 \\
			IPOD \cite{yang2018ipod} &LiDAR+Mono& 84.10 & 76.40 & 75.30 \\
			F-ConvNet\cite{wang2019frustum} &LiDAR+Mono& 89.02 & 78.80 & 77.09 \\ \hline
			VoxelNet\cite{zhou2018voxelnet} &LiDAR& 81.97 & 65.46 & 62.85 \\
			PointPillars \cite{lang2018pointpillars} &LiDAR &87.29  & 76.99 & 70.84  \\
			PointRCNN\cite{shi2019pointrcnn} &LiDAR& 88.88 & 78.63 & 77.38 \\
			SECOND\cite{yan2018second} &LiDAR & 88.15 & 78.33 & 77.25 \\ \hline
			\multicolumn{1}{r}{SECOND+$\mathcal{L}_\text{IoU}$} &LiDAR& \multicolumn{1}{l}{\textbf{89.16}} & \multicolumn{1}{c}{\textit{\textbf{78.99}}} & \multicolumn{1}{c }{\textit{\textbf{77.78}}}\\
			\multicolumn{1}{r}{SECOND+$\mathcal{L}_\text{GIoU}$} &LiDAR& \multicolumn{1}{l}{\textit{\textbf{89.15}}} & \multicolumn{1}{c}{\textbf{79.14}} & \multicolumn{1}{c }{\textbf{78.11}}  \\ 
		\end{tabular}
	}
	\caption{\normalfont Comparison with other public methods on the KITTI validation dataset for 3D ``Car'' detection. For easy understanding, we have highlighted the top two numbers in bold and italic for each column. All the numbers are the higher the better.}
	\label{tab:evaluation_with_other_methods_on_val_kitti}
\end{table}

In the above subsections, we have compared the implemented new loss with $\mathbf{L}_1$ loss based on different baselines. In this subsection, 
We compare the improved baseline with state-of-the-art methods of 3D object detection on both val split and test split of KITTI \cite{geiger2012we} 3D object detection benchmark. First of all, \tabref{tab:evaluation_with_other_methods_on_val_kitti} gives the comparison results on validation dataset. We have listed nearly all the top results with publications here including: multi-modalities fusion-based  \cite{chen2017multi,qi2018frustum,ku2018joint,liang2018deep,yang2018ipod,wang2019frustum}, one-stage- \cite{yan2018second,zhou2018voxelnet,lang2018pointpillars} and two-stage-based \cite{shi2019pointrcnn} approaches. Among all the methods, the improved baseline with $\mathcal{L}_\text{IoU}$ and $\mathcal{L}_\text{GIoU}$ achieved the best results on all the three categories and it even performs much better than other fusion-based and two-stage-based methods. 

\begin{table}[]
	\resizebox{0.45\textwidth}{!}
	{%
		\begin{tabular}{r c ccc}
		\hline
		\multicolumn{1}{c }{\multirow{2}{*}{Methods}}& \multicolumn{1}{c}{\multirow{2}{*}{Modality}} & \multicolumn{3}{c}{\textbf{AP70}} \\
		\multicolumn{1}{c}{}& \multicolumn{1}{c}{} & \multicolumn{1}{l}{Easy} & \multicolumn{1}{l}{Mod} & \multicolumn{1}{l}{Hard}\\ \hline
			MV3D\cite{chen2017multi} &LiDAR+Mono &71.09 & 62.35 & 55.12 \\
			F-PointNet\cite{qi2018frustum} &LiDAR+Mono & 81.20 & 70.29 & 62.19 \\
			AVOD-FPN\cite{ku2018joint} &LiDAR+Mono &81.94 & 71.88 & 66.38 \\
			ContFusion\cite{liang2018deep} &LiDAR+Mono & 82.54 & 66.22 & 64.04 \\
			IPOD \cite{yang2018ipod} &LiDAR+Mono &79.75& 72.57 & 66.33 \\ 
			F-ConvNet\cite{wang2019frustum} &LiDAR+Mono &\textit{\textbf{85.88}} & \textbf{76.51} & 68.08 \\ \hline	
			
			VoxelNet\cite{zhou2018voxelnet} &LiDAR & 77.47 & 65.11 & 57.73 \\
			PointPillars \cite{lang2018pointpillars} &LiDAR &79.05  & 74.99 & \textit{\textbf{68.30}}  \\
			PointRCNN\cite{shi2019pointrcnn} &LiDAR &\textbf{85.94} & 75.76 & \textbf{68.32} \\	
			SECOND\cite{yan2018second} &LiDAR &84.04 & 75.38 & 67.36 \\ \hline
			\multicolumn{1}{r}{SECOND+$\mathcal{L}_\text{IoU}$} &LiDAR & \multicolumn{1}{l}{84.43} & \multicolumn{1}{c}{\textit{\textbf{76.28}}} &\multicolumn{1}{c}{68.22}  \\ \hline
		\end{tabular}
	}
	\caption{\normalfont Comparison with other public methods on the KITTI testing dataset for 3D ``Car'' detection. For easy understanding, we have highlighted the top two numbers in bold and italic for each column. All the numbers are the higher the better.}
	\label{tab:evaluation_with_other_methods_on_test_kitti}
\end{table}
 \tabref{tab:evaluation_with_other_methods_on_test_kitti} gives the evaluation results on the KITTI testing benchmark. We achieved the results on testing split submitting the results on KITTI's online evaluation server and the results of other methods are obtained from their publications respectively. One important thing is that our model submitted to the test server is trained with half-half split as used on the validation dataset rather than using a bigger training split (\eg, \cite{lang2018pointpillars}). From the table, we can find that the proposed loss improved the performance of the baseline \cite{yan2018second} for all the three types. Especially for ``moderate'' and ``hard'' categories, the improvement nearly reaches one point. Furthermore, for the ``moderate'' and ``hard'' types, the proposed loss achieved comparable or even better results with the state-of-the-art sensor fusion-based \cite{wang2019frustum} or two-stages-based methods \cite{shi2019pointrcnn}.

\section{Conclusion and Future Works}
In this paper, we have addressed the 2D/3D object detection problem by introducing the IoU loss for two rotated Bboxes. We proposed a unified framework independent IoU loss layer which can be directly applied for axis-aligned or rotated 2D/3D object detection frameworks. By integrating this IoU loss layer into several state-of-the-art 3D object detectors, consistent improvements have been achieved for both 2D detection in bird-eye-view and 3D object detection in point cloud. Especially, the proposed IoU loss performs much better when the IoU threshold is set at a high value. In the future, we would like to extend the current IoU loss layer to more general 3D object detection cases, \eg, Bboxes with three orientation parameters.   

{\small
\bibliographystyle{unsrt}
\bibliography{references}
}

\end{document}